\documentclass[11pt]{article}

\PassOptionsToPackage{table}{xcolor}
\usepackage[preprint]{acl}

\makeatletter
\@ifpackageloaded{lineno}{%
  \IfFormatAtLeastTF{2025-06-01}{%
    \RemoveFromHook{build/column/before}[lineno]%
    \AddToHook{build/column/after}{%
      \setbox\@outputbox\vbox{%
        \boxmaxdepth\@maxdepth
        \protected@write\@auxout{}{%
          \string\def\string\@LN@column{\if@firstcolumn1\else2\fi}}%
        \box\@outputbox}}%
  }{}%
}{}
\makeatother

\usepackage{times}
\usepackage{latexsym}
\usepackage{fontawesome5}
\usepackage{microtype}

\usepackage[UTF8,scheme=plain,fontset=fandol]{ctex}
\usepackage{subcaption}
\usepackage{amsmath,amssymb}
\usepackage{booktabs}
\usepackage{tabularx}
\usepackage{array}
\usepackage{graphicx}
\usepackage{flafter}
\usepackage[table]{xcolor}
\usepackage{multirow}
\usepackage{needspace}

\usepackage{tcolorbox}
\tcbuselibrary{skins,breakable}

\definecolor{promptnavy}{HTML}{223750}
\definecolor{promptborder}{HTML}{C8D7E7}
\definecolor{promptbadge}{HTML}{EAF1F7}

\newtcolorbox{agentpromptbox}[1]{
  enhanced,
  breakable,
  width=\linewidth,
  colback=white,
  colframe=promptborder,
  coltext=black,
  boxrule=0.65pt,
  arc=3mm,
  boxsep=0pt,
  left=3mm,
  right=3mm,
  top=4mm,
  bottom=3mm,
  before skip=12pt,
  after skip=10pt,
  fontupper=\small,
  fontlower=\small,
  before upper={\setlength{\parindent}{0pt}},
  before lower={\setlength{\parindent}{0pt}},
  title={#1},
  fonttitle=\small\bfseries,
  coltitle=white,
  colbacktitle=promptnavy,
  attach boxed title to top left={
    xshift=3mm,
    yshift=-2mm
  },
  boxed title style={
    colback=promptnavy,
    colframe=promptnavy,
    boxrule=0pt,
    arc=1.5mm,
    boxsep=0pt,
    left=2.5mm,
    right=2.5mm,
    top=1.5mm,
    bottom=1.5mm
  },
  segmentation style={
    draw=promptborder,
    line width=0.5pt,
    dash pattern=on 3pt off 3pt
  }
}

\title{Bridging Foundational and Agentic Capabilities of LLMs}

\author{%
  \textbf{Wei Guo\textsuperscript{1},
  Yuanjian Xu\textsuperscript{2},
  Jianing Hao\textsuperscript{2,3},
  Yongyao Duan\textsuperscript{4},}\\
  \textbf{Ran Chen\textsuperscript{4},
  Zhong Li\textsuperscript{5},
  Guang Zhang\textsuperscript{2,\faEnvelope}}\\[0.4em]
  {\normalfont\small\textsuperscript{1}Nankai University\quad
  \textsuperscript{2}The Hong Kong University of Science and Technology (Guangzhou)}\\
  {\normalfont\small\textsuperscript{3}Ant Group\quad
  \textsuperscript{4}OpenCSG\quad
  \textsuperscript{5}Microsoft Research Asia\quad
  \faEnvelope\ Corresponding author}
}
\begin{document}

\maketitle

% =========================
% Abstract
% =========================

\begin{abstract}
Large language models (LLMs) are increasingly deployed as agents, yet it remains unclear how their foundational capabilities translate into agentic capabilities. End-to-end scores leave unresolved whether low benchmark capability or weak deployment drives a failure, limiting actionable diagnosis. We bridge foundational and agentic capabilities through an operation-level measurement framework. We represent each model with a foundational profile derived from capability benchmarks and an operation profile derived from interaction trajectories, and define capability realization gaps as deviations of observed operation performance from predictions based on foundational capabilities. Experiments on fifteen model services in \mbox{$\tau^2$-Bench} show that foundational profiles only partially predict agentic operations. Models with comparable profiles can differ substantially in execution, and these gaps remain consistent across task subsets. Operation profiles add information and substantially improve task-score prediction. The framework locates execution departures at the operation level to inform model development and agent evaluation. We will release our code and analysis artifacts.
\end{abstract}

% =========================
% Main Text
% =========================

\section{Introduction}
\label{sec:introduction}

Large language models (LLMs) have shown considerable promise as agents that interact with users, invoke tools, and respond to environmental feedback over multi-step tasks. This progress has motivated benchmarks for function calling, interactive tool use, and end-to-end task completion \citep{patil2025bfcl,lu2025toolsandbox,barres2025tau2bench,luo2025mcpuniverse}. Yet end-to-end scores say little about how foundational benchmark dimensions translate into planning, action selection, and recovery. Connecting these levels separates low benchmark profiles from execution that falls below profile-based expectations.

Prior work maps evaluations outside agent environments to aggregate performance \citep{song2026pace} and separately diagnoses intermediate behavior from execution trajectories \citep{ou2025agentdiagnose}. Linking these views at the operation level lets us test how foundational profiles predict specific execution behaviors and how much task-outcome information process measurements add. Aggregate prediction leaves the execution stage of a capability gap unidentified. A useful account must connect offline capability structure to specific interactive behaviors, generalize across model families, and determine whether those behaviors explain outcome variation left unresolved by benchmark scores. These requirements motivate the questions below.

To address this gap, we organize our study around two research questions: \textcolor{blue!70!black}{\textit{\textbf{RQ1.} How well do foundational benchmark scores predict operation scores?}} and \textcolor{purple!70!black}{\textit{\textbf{RQ2.} Do operation scores provide additional information about overall task performance?}} Each model receives a foundational profile from capability benchmarks and an operation profile from trajectory annotations; Section~\ref{sec:framework} defines both. For each provider in turn, we train statistical predictors on the remaining providers and evaluate the excluded provider. Residuals answer RQ1 by locating departures from expected operation scores. The same held-out splits answer RQ2 by comparing foundational, operation, and joint predictors of task performance.

We evaluate model configurations on interaction trajectories from the Banking tasks in $\tau^2$-Bench \citep{barres2025tau2bench}. Foundational profiles explain part of the variation in operation scores, with systematic residuals across models and operations. Operation profiles further improve held-out prediction of overall task performance. Our contributions are a framework linking foundational indicators, execution operations, and task outcomes under controlled conditions; a provider-held-out mapping and residual analysis of execution departures; and evidence that combining foundational and operation profiles improves held-out Banking reward prediction.
\section{Related Work}
\label{sec:related-work}

End-to-end agent benchmarks evaluate function calling, stateful tool use, dual-control conversations, and interaction with real-world MCP servers, and they primarily report task-level success \citep{patil2025bfcl,lu2025toolsandbox,barres2025tau2bench,luo2025mcpuniverse}. Capability- and process-oriented evaluations provide complementary explanations for these outcomes. MMAU separates understanding, reasoning, planning, problem-solving, and self-correction in offline tasks \citep{yin-etal-2025-mmau}, ADeLe places task demands and model abilities on interpretable scales to predict instance-level performance \citep{zhou2026general}, and PACE shows that a small set of non-agentic items can predict aggregate agentic benchmark scores \citep{song2026pace}. These approaches map capabilities to aggregate outcomes. Process-level evaluation examines execution directly: T-Eval scores tool-use components step by step against reference trajectories \citep{chen2024teval}, and AgentDiagnose scores competencies such as task decomposition, observation reading, and self-verification from agent trajectories \citep{ou2025agentdiagnose}. Our design additionally pairs process scores with independently measured foundational profiles for the same models. Related work also documents capability--action gaps in specific settings: strong diagnostic reasoning can coexist with weak evidence collection \citep{long2026strong}, recognizing the need for a tool can still end in a missed call \citep{cheng2026toolnecessity}, and perturbations of intermediate reasoning reveal how much later outputs depend on it \citep{lanham2023faithfulness}. We connect these lines by measuring foundational profiles, trajectory-level operations, and task outcomes under shared task conditions, then testing both capability-to-operation prediction and the incremental outcome information in operation profiles. Table~\ref{tab:related-work} summarizes these differences.

\begin{table}[t]
\centering
\small
\setlength{\tabcolsep}{3.5pt}
\resizebox{\columnwidth}{!}{%
\begin{tabular}{lcccccc}
\toprule
\textbf{Work} & \textbf{Inter.} & \textbf{Found.} & \textbf{Proc.} & \textbf{F$\rightarrow$Y} & \textbf{F$\rightarrow$P} & \textbf{P$\rightarrow$Y} \\
\midrule
BFCL \citep{patil2025bfcl} & \checkmark & $\times$ & $\times$ & $\times$ & $\times$ & $\times$ \\
ToolSandbox \citep{lu2025toolsandbox} & \checkmark & $\times$ & \checkmark & $\times$ & $\times$ & $\times$ \\
$\tau^2$-Bench \citep{barres2025tau2bench} & \checkmark & $\times$ & $\times$ & $\times$ & $\times$ & $\times$ \\
MCP-Universe \citep{luo2025mcpuniverse} & \checkmark & $\times$ & $\times$ & $\times$ & $\times$ & $\times$ \\
\midrule
MMAU \citep{yin-etal-2025-mmau} & $\times$ & \checkmark & $\times$ & $\times$ & $\times$ & $\times$ \\
ADeLe \citep{zhou2026general} & $\times$ & \checkmark & $\times$ & \checkmark & $\times$ & $\times$ \\
PACE \citep{song2026pace} & \checkmark & \checkmark & $\times$ & \checkmark & $\times$ & $\times$ \\
\midrule
T-Eval \citep{chen2024teval} & $\times$ & $\times$ & \checkmark & $\times$ & $\times$ & $\times$ \\
AgentDiagnose \citep{ou2025agentdiagnose} & \checkmark & $\times$ & \checkmark & $\times$ & $\times$ & $\times$ \\
\midrule
\textbf{Ours} & \checkmark & \checkmark & \checkmark & \checkmark & \checkmark & \checkmark \\
\bottomrule
\end{tabular}%
}
\caption{\textbf{Coverage of related agent and capability evaluations.}
Inter. denotes multi-turn interactive evaluation, Found. a separate
foundational capability profile, and Proc. trajectory-level process
measurement. $F$, $P$, and $Y$ denote foundational profiles, process
measurements, and task outcomes; arrows indicate which relationship a
study tests. \checkmark{} denotes support and $\times$ no support.}
\label{tab:related-work}
\end{table}

\section{Framework}
\label{sec:framework}

\begin{figure}[t]
\centering
\includegraphics[width=\columnwidth]{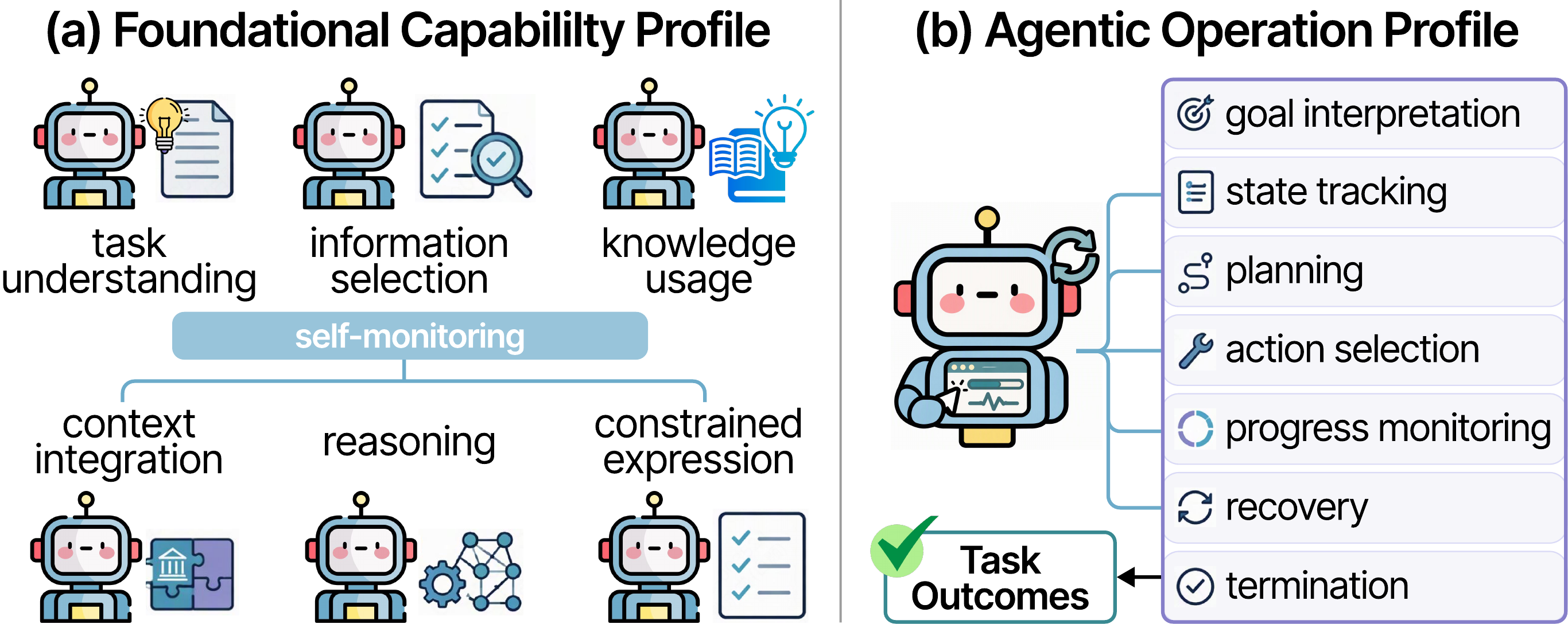}
\caption{\textbf{Framework linking foundational capabilities, agentic
operations, and task outcomes.} Foundational capabilities describe
resources measured outside the agent environment; agentic operations
describe how a model deploys them during interaction. We measure the
relationship between the two profiles and test whether operation profiles
add information about task outcomes.}
\label{fig:framework-overview}
\end{figure}

We study a resource--deployment--outcome sequence. For model $m$, the
foundational profile $F_m=\Theta_m$ summarizes resources measured before
interaction, the operation profile $O_m$ records their deployment along a
trajectory, and $Y_m$ is the resulting task outcome. Each service uses a
fixed inference protocol and a common action interface. The task fixes the
horizon, observability, action space, and feedback. Figure~\ref{fig:framework-overview}
summarizes this measurement structure.

The section follows the same logic: Section~\ref{sec:foundational-profile}
defines the available resources, Section~\ref{sec:operation-profile}
measures their deployment in trajectories, and
Section~\ref{sec:linking-profiles} links both profiles to outcomes and
controlled execution tests.

\subsection{Foundational Capability Profile}
\label{sec:foundational-profile}

Following existing capability taxonomies
\citep{yin-etal-2025-mmau,mo-etal-2025-cdt,zhou2026general}, we define
seven dimensions. Task Understanding (TU) recovers the objective, success
conditions, and relevant constraints. Information Selection (IS)
identifies evidence relevant to the current decision while excluding
distractors. Knowledge Use (KU) retrieves applicable parametric knowledge
or applies reliable knowledge supplied in context. Context Integration
(CI) combines the selected evidence, knowledge, prior context, and
intermediate results into a consistent representation. Reasoning (RE)
derives conclusions from that representation. Constrained Expression
(CE) realizes the conclusion while satisfying content, format, and schema
requirements. Self-Monitoring (SM) detects uncertainty, contradictions,
insufficient evidence, and candidate errors.

The dimensions follow a coherent processing sequence: the model
understands the task, prepares relevant information and knowledge,
integrates the resulting inputs, reasons over them, and expresses the
result. Self-Monitoring provides cross-stage feedback and may trigger
revision of integration, reasoning, or expression. For model $m$,
$\Theta_m$ denotes the resulting seven-score foundational profile. We
report the seven coordinates separately to preserve capability-specific
variation across the heterogeneous source benchmarks.

\subsection{Agentic Operation Profile}
\label{sec:operation-profile}

Agentic operations describe what the model does with these resources in a
particular trajectory. Goal Interpretation (GI) grounds subsequent
decisions in the user's objective and constraints. State Tracking (ST)
updates the task state after new information or environment feedback.
Planning (PL) orders the steps required to reach the goal. Action
Selection (AS) chooses the next tool call or user-facing action and its
arguments. Progress Monitoring (PM) identifies what is complete, pending,
failed, or blocked. Recovery (RC) repairs a failed action or an erroneous
decision. Termination (TM) decides whether to continue, stop, or transfer.
The seven operations form a recurring control loop. State Tracking and
Progress Monitoring recur after informative transitions, while Recovery
returns control to Planning or Action Selection. Their relation to
foundational capabilities is many-to-many: one capability can support
several operations, and one operation can draw on several capabilities.

Operations are measured only when a trajectory provides enough evidence
to judge them. Before review, a task contract records the goal,
constraints, required actions, and valid terminal states.
Table~\ref{tab:operation-annotation-rules} lists the evidence used for
each operation. All operations share one label rule: 0 for a failed or
omitted applicable requirement, 0.5 for partial satisfaction without a
failure, and 1 for complete satisfaction. Recovery additionally requires a
preceding error and an opportunity to respond to it. A trajectory without
an observable opportunity is recorded as NA and excluded from aggregation. Within each
trajectory, labels are averaged by operation and scaled to 0--100; a
model's operation score averages over trajectories with a scoring
opportunity. For model $m$, $O_m$ denotes the seven-score operation
profile and $A_m$ its overall operation score, which first averages
applicable operations within each trajectory and then averages across
trajectories.

\begin{table}[t]
\centering
\small
\setlength{\tabcolsep}{4pt}
\renewcommand{\arraystretch}{1.08}
\begin{tabularx}{\columnwidth}{@{}l>{\raggedright\arraybackslash}X@{}}
\toprule
\textbf{Op.} & \textbf{Scoring evidence} \\
\midrule
GI & Later decisions preserve the user's goal, target entities, constraints, and success conditions. \\
ST & New information and tool results are incorporated into the current state and used correctly. \\
PL & The action sequence respects dependencies and task constraints. \\
AS & Actions and arguments are appropriate, authorized, and consistent with the applicable rules. \\
PM & Completed, pending, failed, and blocked requirements are distinguished correctly. \\
RC & A preceding error is corrected, and the correction is retained in later execution. \\
TM & Continuing, stopping, or transferring is consistent with the remaining requirements and final state. \\
\bottomrule
\end{tabularx}
\caption{\textbf{Trajectory evidence used to score the seven agentic
operations.} Each row states the behavioral evidence required for one
operation; the common 0/0.5/1 label rule and aggregation procedure are
defined in Section~\ref{sec:operation-profile}.}
\label{tab:operation-annotation-rules}
\end{table}

\subsection{Analysis Methodology}
\label{sec:linking-profiles}
\label{sec:association-prediction}

\paragraph{Pairwise associations.}
Let $F_m=\Theta_m$ and $O_m$ denote the foundational and operation
profiles of model $m$. We compute the $7\times7$ Pearson correlation
matrix between their dimensions and use Spearman correlation as a
rank-based robustness check. Permutation tests, multiple-comparison
correction, and family-composition analyses quantify uncertainty and
sensitivity. We use these associations as descriptive summaries. The
paired intervention design below evaluates effects on downstream
execution.

\paragraph{Family-held-out prediction.}
Models from the same family may share architectures, training data, and
post-training procedures. We therefore hold out one complete family at a
time. Let $\mathcal{I}_{\mathrm{train}}$ be the indices of models in the
current outer training fold. For each target, we fit a ridge predictor with
an unpenalized intercept:
\begin{equation}
\min_{b,\beta}
\sum_{i\in\mathcal{I}_{\mathrm{train}}}
\left(y_i-b-z_i^\top\beta\right)^2
+\alpha\lVert\beta\rVert_2^2,
\label{eq:ridge-objective}
\end{equation}
where $y_i$ is the scalar target for training model $i$, $z_i$ is its
standardized feature vector, $b$ is the intercept, $\beta$ is the
coefficient vector, and $\alpha$ is the ridge penalty. Feature
standardization is estimated from $\mathcal{I}_{\mathrm{train}}$ only.

\paragraph{Residual diagnostics.}
Let $f(m)$ denote the family of held-out model $m$. We define the
operation-level and overall capability realization gaps as
\begin{equation}
\begin{aligned}
r_{m,o} &=
O_{m,o}-\widehat O_{m,o}^{(-f(m))},\\
\Delta_m^{\mathrm{op}} &=
A_m-\widehat A_m^{(-f(m))},
\end{aligned}
\label{eq:realization-gaps}
\end{equation}
where the superscript $(-f(m))$ indicates that the predictor was trained
without family $f(m)$. Positive values indicate execution above the
prediction from the foundational profile; negative values indicate
execution below it. We interpret a residual only when its target has useful
held-out predictive skill.

\paragraph{Outcome prediction.}
For the scalar task outcome $Y_m$, we compare a training-mean baseline with
three learned predictors using $F$, $O$, or $F+O$. All learned predictors
use the same family splits, estimator class, and preprocessing.
Improvement of the joint predictor over the foundational-only predictor
quantifies the incremental outcome information in operation profiles.
Because operation scores and outcomes share trajectories, this analysis
measures within-corpus prediction on the Banking trajectories.

\paragraph{Evaluation metrics.}
We primarily evaluate pooled held-out predictions using Spearman
correlation and mean absolute error. Kendall's $\tau_b$, root mean squared
error, and predictive skill relative to the fold-specific training mean
serve as complementary metrics. Let $\mathcal{M}$ denote the evaluated
models, $t_m$ the target for model $m$, and
$\overline t_{\mathrm{train}(m)}$ the mean target in its outer training
fold. Predictive skill is
\begin{equation}
Q^2(t)
=
1-
\frac{\sum_{m\in\mathcal{M}}(t_m-\widehat t_m^{(-f(m))})^2}
{\sum_{m\in\mathcal{M}}(t_m-\overline t_{\mathrm{train}(m)})^2}.
\label{eq:heldout-predictive-skill}
\end{equation}
Because $Q^2$ uses held-out predictions, it may take negative values.

\paragraph{Shared-initial-state experiments.}
The shared-initial-state experiments compare repeated continuations
from the same dialogue history, environment state, and available tools.
Natural continuation and a generic checking prompt provide two
conditions. Scoring separates violations, completed compliant
operations, and unresolved behaviors. Single-step and short
continuations record whether subsequent actions preserve compliance.
The complete protocol appears in Appendix~\ref{app:shared-state-protocol}.

\paragraph{Counterfactual interventions.}
\label{sec:counterfactual-interventions}
\label{sec:planning-intervention}
To test whether operation-relevant content affects later execution, we
compare paired runs that modify this content while holding the task
state, tools, goal, and budget fixed. Damage interventions minimally
corrupt valid content; repair interventions minimally correct invalid
but repairable content. Semantics-preserving rewrites control for
surface-form changes. Repeated runs are aggregated within the base task,
which is also the unit of statistical resampling.

\section{Experiments}
\label{sec:experiments}

We first map foundational profiles to operation profiles and assess the
task-score information added by operation measurements. Cross-task splits
and controlled continuations then probe whether capability realization gaps
persist across tasks and reappear in execution.

\subsection{Setup}
\label{sec:experimental-setup}

The study covers the model services in Table~\ref{tab:model-profiles},
spanning eight model families.\footnote{The original service identifiers
retain the following snapshot tags: DeepSeek V4 Pro (0813), DeepSeek V4
Flash (0731), Qwen3.8-Max (20260803), and MiniMax M3 (20260531). We omit
these tags from model names in the text, tables, and figures for
readability.} Foundational profiles combine public evaluation results with
local measurements on a common scale. Table~\ref{tab:foundational-battery}
maps each dimension to its evaluation and score. For Self-Monitoring, the
model answers a question and then estimates the probability that its answer
is correct. The score averages the own-answer AUROCs from the two evaluation
sets. Appendix~\ref{app:foundational-evaluation-config} documents effective
cohorts and run configurations.

\begin{table*}[t]
\centering
\small
\setlength{\tabcolsep}{5pt}
\renewcommand{\arraystretch}{1.08}
\begin{tabularx}{\textwidth}{@{}l>{\raggedright\arraybackslash}p{0.36\textwidth}lX@{}}
\toprule
\textbf{Dimension} & \textbf{Evaluation} & \textbf{Reference size} & \textbf{Primary score} \\
\midrule
TU & MultiChallenge \citep{sirdeshmukh2025multichallenge} & Source-specific & Overall rubric pass rate \\
IS & MRCR, 8 needles, up to 128K \citep{vodrahalli2024michelangelo,openai2025mrcr} & 484 & Mean sequence-match score \\
KU & MMLU-Pro \citep{wang2024mmlupro} & 12,032 & Answer accuracy \\
CI & LongBench v2 \citep{bai2025longbenchv2} & 503 & Answer accuracy \\
RE & GPQA Diamond and text-only HLE, without tools & 198 + 2,158 & Equal-weight mean accuracy \\
SM & GPQA Diamond and SimpleQA \citep{rein2023gpqa,openai2024simpleqa} & 198 + 300 & Equal-weight mean confidence AUROC \\
CE & IFBench \citep{pyatkin2025ifbench} & 300 & Prompt-level loose accuracy \\
\bottomrule
\end{tabularx}
\caption{\textbf{Evaluation suite used to construct the foundational
capability profile.} Each row maps one profile dimension to its
evaluation, reference size, and primary score. Effective denominators
for source-specific scores are recorded in the supplementary materials;
all primary scores are reported on a 0--100 scale.}
\label{tab:foundational-battery}
\end{table*}

Agentic operation profiles use the full Banking trajectory corpus,
distributed nearly evenly across services. The tasks combine service goals with
identity, confirmation, and policy constraints. We score each operation on
complete trajectories and use the environment's native reward
as the task outcome. Section~\ref{sec:operation-profile} defines scoring and
aggregation; Appendix~\ref{app:trajectory-annotation-protocol} gives the
annotation protocol.

Prediction uses the model as the unit of analysis and holds out complete
model families. Ridge predictors standardize features within each training
fold and use the common penalty $\alpha=10$. Appendix~\ref{app:alpha10-reference}
reports neighboring penalties and the full metric suite. The main text
emphasizes held-out error, rank agreement, and predictive skill relative to
the training-mean baseline.

Complete-family holdout prevents closely related services from appearing in
both training and evaluation. We reuse the folds and keep the estimator and
preprocessing fixed across feature sets, so performance differences reflect
the information in each profile. Operation targets include only trajectories
with a scoring opportunity.

\providecommand{\analysisplaceholder}[3]{%
  \fbox{\parbox[c][#2][c]{\dimexpr#1-2\fboxsep-2\fboxrule\relax}{%
    \centering\small #3}}}

\newcommand{\resultbox}[2]{%
  \par
  \noindent\begingroup
  \setlength{\fboxsep}{5pt}%
  \fcolorbox{blue!45!black}{blue!4}{%
    \parbox{\dimexpr\linewidth-2\fboxsep-2\fboxrule\relax}{%
      \small
      \textcolor{blue!65!black}{\ensuremath{\blacktriangleright}}\hspace{0.45em}%
      \textbf{#1.} #2}}%
  \endgroup
  \par\vspace{-0.2\baselineskip}
}

\subsection{Main Results}
\label{sec:results}

\Needspace{7\baselineskip}
\resultbox{Takeaway}{Foundational and operation profiles reveal different model strengths.}

\paragraph{Foundational Capability and Operation Profiles}
\label{sec:results-profiles}
Foundational and operation profiles rank models differently,
distinguishing capabilities measured in benchmarks from performance
during task execution. Table~\ref{tab:model-profiles} pairs each model's
seven-dimensional foundational profile with its seven-dimensional
operation profile. Operation scores identify strengths in Goal
Interpretation and State Tracking, alongside execution differences in
Action Selection, Recovery, and Termination. The two profiles separately
characterize the capabilities a model possesses and how it applies them
during interaction.

Dimension-level comparison prevents a strong average from masking a local
weakness. High knowledge or reasoning can coexist with errors in state
updates, action choice, or recovery; consistent execution can also offset a
less prominent benchmark profile. The predictive mapping below turns this
separation into an expected operation profile and a deployment gap.

\newcommand{\modelicon}[1]{%
  \raisebox{-0.18\height}{\includegraphics[height=1.15em]{fig/provider_icons/#1.png}}%
  \hspace{0.35em}}
\definecolor{profileblue}{RGB}{234,246,252}

\begin{table*}[!t]
\centering
\definecolor{profilehigh}{RGB}{94,170,165}
\footnotesize
\setlength{\tabcolsep}{3pt}
\renewcommand{\arraystretch}{1.12}
\begin{tabularx}{\textwidth}{@{}>{\raggedright\arraybackslash}Xrrrrrrr@{\hspace{9pt}}rrrrrrr@{}}
\toprule
\multicolumn{1}{c}{} &
\multicolumn{7}{c}{\textbf{Foundational Capability Profile}} &
\multicolumn{7}{c}{\textbf{Agentic Operation Profile}} \\
\cmidrule(lr){2-8}\cmidrule(lr){9-15}
\textbf{Model} &
\textbf{TU} & \textbf{IS} & \textbf{KU} & \textbf{CI} &
\textbf{RE} & \textbf{SM} & \textbf{CE} &
\textbf{GI} & \textbf{ST} & \textbf{PL} & \textbf{AS} &
\textbf{PM} & \textbf{RC} & \textbf{TM} \\
\midrule
\modelicon{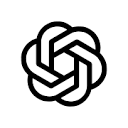}GPT-5.6 Sol
& \cellcolor{profilehigh!43}73.0 & \cellcolor{profilehigh!45}97.6 & \cellcolor{profilehigh!30}89.1 & \cellcolor{profilehigh!34}67.1 & \cellcolor{profilehigh!36}71.8 & \cellcolor{profilehigh!5}67.8 & \cellcolor{profilehigh!21}72.7
& \cellcolor{profilehigh!40}79.2 & \cellcolor{profilehigh!27}74.5 & \cellcolor{profilehigh!33}79.3 & \cellcolor{profilehigh!24}47.4 & \cellcolor{profilehigh!29}72.7 & \cellcolor{profilehigh!33}60.5 & \cellcolor{profilehigh!23}40.1 \\
\modelicon{openai}GPT-5.6 Luna
& \cellcolor{profilehigh!39}68.7 & \cellcolor{profilehigh!33}81.8 & \cellcolor{profilehigh!11}86.0 & \cellcolor{profilehigh!27}62.4 & \cellcolor{profilehigh!9}65.3 & \cellcolor{profilehigh!8}69.1 & \cellcolor{profilehigh!2}64.3
& \cellcolor{profilehigh!38}78.3 & \cellcolor{profilehigh!27}74.6 & \cellcolor{profilehigh!27}76.8 & \cellcolor{profilehigh!37}55.2 & \cellcolor{profilehigh!35}76.7 & \cellcolor{profilehigh!20}51.1 & \cellcolor{profilehigh!45}\textbf{49.7} \\
\modelicon{openai}GPT-5.5
& \cellcolor{profilehigh!43}73.3 & \cellcolor{profilehigh!42}94.2 & \cellcolor{profilehigh!24}88.1 & \cellcolor{profilehigh!34}67.0 & \cellcolor{profilehigh!27}69.7 & \cellcolor{profilehigh!5}67.5 & \cellcolor{profilehigh!29}75.9
& \cellcolor{profilehigh!35}77.1 & \cellcolor{profilehigh!28}75.3 & \cellcolor{profilehigh!45}84.1 & \cellcolor{profilehigh!35}54.1 & \cellcolor{profilehigh!29}72.7 & \cellcolor{profilehigh!2}37.9 & \cellcolor{profilehigh!34}44.7 \\
\arrayrulecolor{black!35}\midrule[0.3pt]\arrayrulecolor{black}
\modelicon{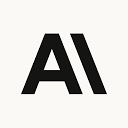}Claude Opus 5
& \cellcolor{profilehigh!42}71.7 & \cellcolor{profilehigh!45}\textbf{97.7} & \cellcolor{profilehigh!45}\textbf{91.6} & \cellcolor{profilehigh!45}\textbf{74.4} & \cellcolor{profilehigh!45}\textbf{74.0} & \cellcolor{profilehigh!21}75.1 & \cellcolor{profilehigh!0}63.4
& \cellcolor{profilehigh!45}\textbf{81.1} & \cellcolor{profilehigh!41}80.7 & \cellcolor{profilehigh!45}\textbf{84.3} & \cellcolor{profilehigh!45}\textbf{59.6} & \cellcolor{profilehigh!45}\textbf{83.3} & \cellcolor{profilehigh!45}\textbf{69.4} & \cellcolor{profilehigh!41}47.9 \\
\modelicon{anthropic}Claude Fable 5
& \cellcolor{profilehigh!45}\textbf{75.3} & \cellcolor{profilehigh!44}96.4 & \cellcolor{profilehigh!44}91.5 & \cellcolor{profilehigh!32}65.2 & \cellcolor{profilehigh!45}\textbf{74.0} & \cellcolor{profilehigh!45}\textbf{86.2} & \cellcolor{profilehigh!0}63.5
& \cellcolor{profilehigh!45}80.9 & \cellcolor{profilehigh!45}\textbf{82.7} & \cellcolor{profilehigh!34}79.5 & \cellcolor{profilehigh!36}54.4 & \cellcolor{profilehigh!38}78.4 & \cellcolor{profilehigh!43}68.3 & \cellcolor{profilehigh!41}47.9 \\
\modelicon{anthropic}Claude Sonnet 5
& \cellcolor{profilehigh!36}66.1 & \cellcolor{profilehigh!31}79.5 & \cellcolor{profilehigh!20}87.5 & \cellcolor{profilehigh!25}61.0 & \cellcolor{profilehigh!13}66.2 & \cellcolor{profilehigh!6}68.3 & \cellcolor{profilehigh!11}68.2
& \cellcolor{profilehigh!45}81.0 & \cellcolor{profilehigh!30}75.9 & \cellcolor{profilehigh!39}81.6 & \cellcolor{profilehigh!43}58.4 & \cellcolor{profilehigh!33}75.3 & \cellcolor{profilehigh!30}58.3 & \cellcolor{profilehigh!42}48.2 \\
\arrayrulecolor{black!35}\midrule[0.3pt]\arrayrulecolor{black}
\modelicon{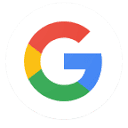}Gemini 3.1 Pro
& \cellcolor{profilehigh!41}71.4 & \cellcolor{profilehigh!35}84.7 & \cellcolor{profilehigh!41}91.0 & \cellcolor{profilehigh!39}70.3 & \cellcolor{profilehigh!31}70.6 & \cellcolor{profilehigh!33}80.4 & \cellcolor{profilehigh!31}77.1
& \cellcolor{profilehigh!1}63.1 & \cellcolor{profilehigh!14}69.0 & \cellcolor{profilehigh!12}70.4 & \cellcolor{profilehigh!10}39.7 & \cellcolor{profilehigh!11}60.4 & \cellcolor{profilehigh!6}40.7 & \cellcolor{profilehigh!4}31.8 \\
\modelicon{google}Gemini 3.6 Flash
& \cellcolor{profilehigh!42}71.7 & \cellcolor{profilehigh!38}88.8 & \cellcolor{profilehigh!31}89.3 & \cellcolor{profilehigh!34}67.0 & \cellcolor{profilehigh!16}66.8 & \cellcolor{profilehigh!0}65.3 & \cellcolor{profilehigh!32}77.3
& \cellcolor{profilehigh!30}75.1 & \cellcolor{profilehigh!33}77.2 & \cellcolor{profilehigh!32}78.7 & \cellcolor{profilehigh!32}52.2 & \cellcolor{profilehigh!28}72.1 & \cellcolor{profilehigh!21}51.7 & \cellcolor{profilehigh!32}43.9 \\
\arrayrulecolor{black!35}\midrule[0.3pt]\arrayrulecolor{black}
\modelicon{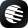}Kimi K3
& \cellcolor{profilehigh!26}54.3 & \cellcolor{profilehigh!25}71.7 & \cellcolor{profilehigh!23}88.0 & \cellcolor{profilehigh!31}64.8 & \cellcolor{profilehigh!23}68.5 & \cellcolor{profilehigh!2}66.4 & \cellcolor{profilehigh!25}74.4
& \cellcolor{profilehigh!28}74.0 & \cellcolor{profilehigh!19}71.3 & \cellcolor{profilehigh!13}70.8 & \cellcolor{profilehigh!24}47.9 & \cellcolor{profilehigh!33}75.0 & \cellcolor{profilehigh!8}42.4 & \cellcolor{profilehigh!30}43.0 \\
\modelicon{moonshot}Kimi K2.6
& \cellcolor{profilehigh!34}63.1 & \cellcolor{profilehigh!20}64.6 & \cellcolor{profilehigh!24}88.1 & \cellcolor{profilehigh!0}43.7 & \cellcolor{profilehigh!2}63.5 & \cellcolor{profilehigh!1}65.6 & \cellcolor{profilehigh!24}73.7
& \cellcolor{profilehigh!19}70.6 & \cellcolor{profilehigh!0}62.7 & \cellcolor{profilehigh!17}72.4 & \cellcolor{profilehigh!17}43.4 & \cellcolor{profilehigh!17}64.4 & \cellcolor{profilehigh!15}47.3 & \cellcolor{profilehigh!15}36.6 \\
\arrayrulecolor{black!35}\midrule[0.3pt]\arrayrulecolor{black}
\modelicon{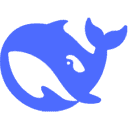}DeepSeek V4 Pro
& \cellcolor{profilehigh!35}64.1 & \cellcolor{profilehigh!34}83.5 & \cellcolor{profilehigh!20}87.5 & \cellcolor{profilehigh!27}62.1 & \cellcolor{profilehigh!14}66.4 & \cellcolor{profilehigh!7}68.4 & \cellcolor{profilehigh!36}79.1
& \cellcolor{profilehigh!38}78.1 & \cellcolor{profilehigh!34}77.6 & \cellcolor{profilehigh!34}79.8 & \cellcolor{profilehigh!24}47.6 & \cellcolor{profilehigh!34}76.2 & \cellcolor{profilehigh!14}46.9 & \cellcolor{profilehigh!38}46.8 \\
\modelicon{deepseek}DeepSeek V4 Flash
& \cellcolor{profilehigh!34}63.5 & \cellcolor{profilehigh!30}78.7 & \cellcolor{profilehigh!12}86.2 & \cellcolor{profilehigh!19}57.0 & \cellcolor{profilehigh!0}63.0 & \cellcolor{profilehigh!3}66.7 & \cellcolor{profilehigh!43}82.0
& \cellcolor{profilehigh!10}66.9 & \cellcolor{profilehigh!8}66.4 & \cellcolor{profilehigh!19}73.2 & \cellcolor{profilehigh!9}38.9 & \cellcolor{profilehigh!12}61.1 & \cellcolor{profilehigh!14}46.6 & \cellcolor{profilehigh!13}35.9 \\
\arrayrulecolor{black!35}\midrule[0.3pt]\arrayrulecolor{black}
\modelicon{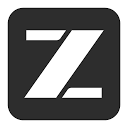}GLM-5.2
& \cellcolor{profilehigh!41}71.3 & \cellcolor{profilehigh!25}72.3 & \cellcolor{profilehigh!15}86.7 & \cellcolor{profilehigh!30}64.0 & \cellcolor{profilehigh!12}65.9 & \cellcolor{profilehigh!12}71.0 & \cellcolor{profilehigh!23}73.3
& \cellcolor{profilehigh!39}78.6 & \cellcolor{profilehigh!37}79.1 & \cellcolor{profilehigh!43}83.4 & \cellcolor{profilehigh!31}51.4 & \cellcolor{profilehigh!37}77.7 & \cellcolor{profilehigh!16}47.9 & \cellcolor{profilehigh!42}48.6 \\
\arrayrulecolor{black!35}\midrule[0.3pt]\arrayrulecolor{black}
\modelicon{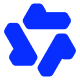}Qwen3.8-Max
& \cellcolor{profilehigh!40}69.8 & \cellcolor{profilehigh!41}92.9 & \cellcolor{profilehigh!27}88.6 & \cellcolor{profilehigh!33}66.3 & \cellcolor{profilehigh!21}68.1 & \cellcolor{profilehigh!14}71.8 & \cellcolor{profilehigh!45}82.8
& \cellcolor{profilehigh!39}78.5 & \cellcolor{profilehigh!28}75.1 & \cellcolor{profilehigh!37}80.8 & \cellcolor{profilehigh!36}54.7 & \cellcolor{profilehigh!39}79.0 & \cellcolor{profilehigh!15}47.5 & \cellcolor{profilehigh!42}48.3 \\
\arrayrulecolor{black!35}\midrule[0.3pt]\arrayrulecolor{black}
\modelicon{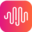}MiniMax M3
& \cellcolor{profilehigh!0}26.8 & \cellcolor{profilehigh!0}39.2 & \cellcolor{profilehigh!0}84.2 & \cellcolor{profilehigh!3}45.6 & \cellcolor{profilehigh!8}65.0 & \cellcolor{profilehigh!11}70.3 & \cellcolor{profilehigh!45}\textbf{83.0}
& \cellcolor{profilehigh!0}62.6 & \cellcolor{profilehigh!2}63.7 & \cellcolor{profilehigh!0}65.2 & \cellcolor{profilehigh!0}34.0 & \cellcolor{profilehigh!0}53.0 & \cellcolor{profilehigh!0}36.4 & \cellcolor{profilehigh!0}30.0 \\
\bottomrule
\end{tabularx}
\caption{\textbf{Foundational capability and agentic operation profiles
for all 15 model services.} Scores use a 0--100 scale; bold marks the
best value in each column, and cell shading runs from white (lower) to
light teal (higher) within each column. Provider icons and separators
identify model families. Dimension definitions appear in
Sections~\ref{sec:foundational-profile} and~\ref{sec:operation-profile};
score provenance and evaluation configurations appear in the appendix.}
\label{tab:model-profiles}
\end{table*}

\Needspace{7\baselineskip}
\resultbox{Takeaway}{Foundational profiles predict each operation category; residuals identify over- and under-deployment.}

\begin{figure*}[!t]
\centering
\includegraphics[width=0.96\textwidth]{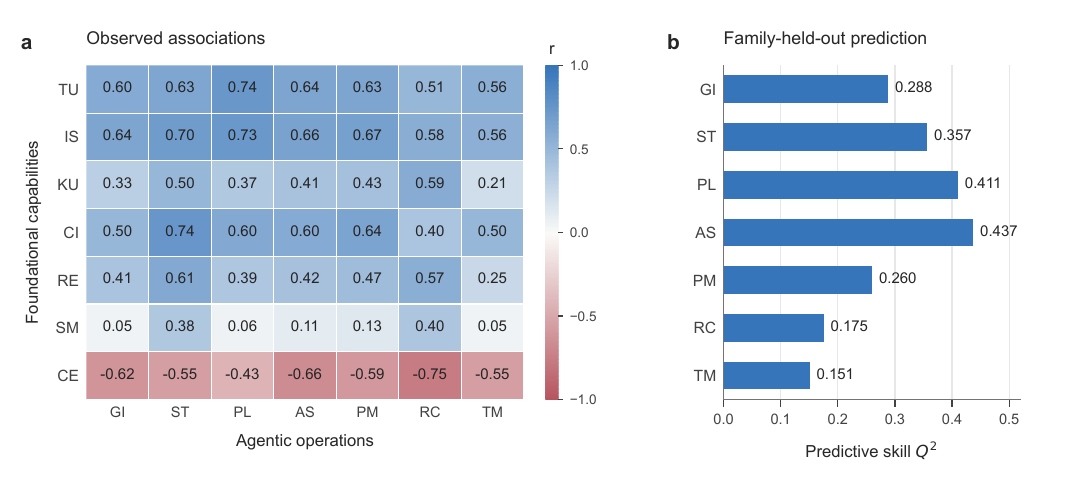}
\caption{\textbf{Mapping foundational capabilities to agentic
operations.} (a) Pearson correlations between the seven foundational and
seven operation scores across 15 models. (b) Model-family-held-out
predictive skill $Q^2$ for each operation using foundational scores and
the common ridge penalty $\alpha=10$. Action Selection and Planning have
the strongest held-out signal, while Recovery and Termination are the
least predictable from foundational scores.}
\label{fig:capability-operation-mapping}
\end{figure*}

\paragraph{Foundational Capability Mapping and Capability Realization Gaps}
\label{sec:results-mapping-gaps}
Foundational profiles have positive model-family-held-out $Q^2$ for
all seven operation categories and improve prediction of the overall
operation score relative to the training-mean baseline.
Figure~\ref{fig:capability-operation-mapping} reports the associations
and held-out prediction results. Capability realization gaps then organize
departures from those predictions. Table~\ref{tab:overall-operation-gap}
shows a clear family pattern: the Claude services tend to exceed their
predicted operation scores, Gemini 3.1 Pro falls below its predictions,
and the remaining services have mixed profiles. The operation-level view
further isolates deviations in specific control functions such as Recovery
and Progress Monitoring. These structured differences point to
model-specific patterns in capability deployment. Appendix
Figure~\ref{fig:operation-residuals} reports every operation-level gap.

The macro gap identifies a broad departure from expected execution; the
operation vector locates its source. Consistent signs indicate a general
deployment pattern; mixed signs isolate stages of the control loop.
This distinction keeps aggregate profiles from obscuring shortfalls in state
maintenance, action choice, monitoring, or recovery.

The family-level ordering becomes more informative when paired with
operation gaps. A positive macro gap can contain local weaknesses, while a
negative macro gap can preserve strengths in individual control functions.
These contrasts locate the responsible execution stage.

\begin{table}[!t]
\centering
\footnotesize
\setlength{\tabcolsep}{2.2pt}
\renewcommand{\arraystretch}{1.04}
\begin{tabularx}{\columnwidth}{@{}>{\raggedright\arraybackslash}Xrrr@{}}
\toprule
\textbf{Model} & \textbf{Observed} & \textbf{Predicted} & \textbf{Gap} \\
\midrule
\modelicon{openai}GPT-5.6 Sol & 65.1 & 68.34 & $-3.24$ \\
\modelicon{openai}GPT-5.6 Luna & 67.6 & 66.97 & $+0.63$ \\
\modelicon{openai}GPT-5.5 & 66.5 & 67.00 & $-0.50$ \\
\arrayrulecolor{black!35}\midrule[0.3pt]\arrayrulecolor{black}
\modelicon{anthropic}Claude Opus 5 & 72.8 & 66.47 & $+6.33$ \\
\modelicon{anthropic}Claude Fable 5 & 70.7 & 62.67 & $+8.03$ \\
\modelicon{anthropic}Claude Sonnet 5 & 69.7 & 64.23 & $+5.47$ \\
\arrayrulecolor{black!35}\midrule[0.3pt]\arrayrulecolor{black}
\modelicon{google}Gemini 3.1 Pro & 54.8 & 68.54 & $-13.74$ \\
\modelicon{google}Gemini 3.6 Flash & 65.8 & 66.31 & $-0.51$ \\
\arrayrulecolor{black!35}\midrule[0.3pt]\arrayrulecolor{black}
\modelicon{moonshot}Kimi K3 & 62.8 & 63.24 & $-0.44$ \\
\modelicon{moonshot}Kimi K2.6 & 58.2 & 61.13 & $-2.93$ \\
\arrayrulecolor{black!35}\midrule[0.3pt]\arrayrulecolor{black}
\modelicon{deepseek}DeepSeek V4 Pro & 66.7 & 63.34 & $+3.36$ \\
\modelicon{deepseek}DeepSeek V4 Flash & 56.0 & 61.73 & $-5.73$ \\
\arrayrulecolor{black!35}\midrule[0.3pt]\arrayrulecolor{black}
\modelicon{zai}GLM-5.2 & 68.5 & 63.44 & $+5.06$ \\
\arrayrulecolor{black!35}\midrule[0.3pt]\arrayrulecolor{black}
\modelicon{qwen}Qwen3.8-Max & 68.9 & 62.33 & $+6.57$ \\
\arrayrulecolor{black!35}\midrule[0.3pt]\arrayrulecolor{black}
\modelicon{minimax}MiniMax M3 & 49.9 & 54.36 & $-4.46$ \\
\bottomrule
\end{tabularx}
\caption{\textbf{Overall capability realization gaps by model.}
Observed and model-family-held-out predicted operation macros use a
0--100 scale under the common penalty $\alpha=10$; the gap is observed
minus predicted. Provider icons and separators identify model families.
Macro-prediction MAE is 4.47 versus 5.92 for the training-mean baseline,
with $Q^2=0.373$.}
\label{tab:overall-operation-gap}
\end{table}

\Needspace{7\baselineskip}
\resultbox{Takeaway}{The joint profile lowers held-out task-score error relative to foundational features alone.}

\begin{figure*}[!t]
\centering
\includegraphics[width=0.96\textwidth]{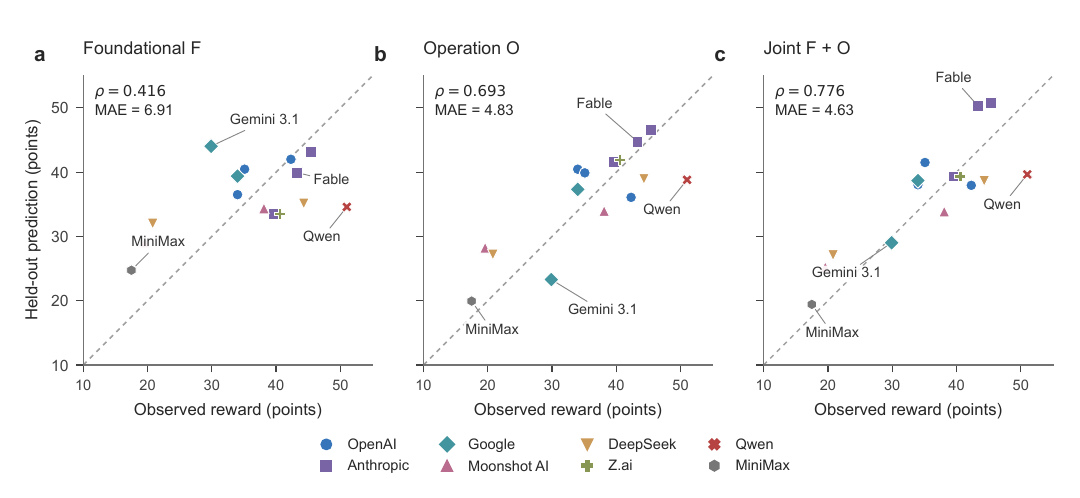}
\caption{\textbf{Model-family-held-out prediction of Banking task
reward from foundational and operation profiles.} Panels compare
foundational features ($F$), operation features ($O$), and their
combination ($F+O$). Each point is one model service, dashed lines mark
equality between predicted and observed reward, and all predictors use
$\alpha=10$.}
\label{fig:banking-reward-prediction}
\end{figure*}

\paragraph{Banking Task-Score Prediction}
\label{sec:results-banking-reward}
The joint profile improves task-score prediction over the foundational
profile under the same model-family-held-out splits. We compare the
foundational profile $F$, operation profile $O$, and joint profile $F+O$,
using each model's mean Banking reward as the target. The mean reward of
the training models provides the baseline. Adding the operation profile
gives the joint predictor the lowest error and strongest rank agreement.
Figure~\ref{fig:banking-reward-prediction} shows the held-out predictions,
and Table~\ref{tab:banking-reward-prediction} reports the full metrics.
The joint gain comes from adding observed deployment signals to
foundational capability scores.

\begin{table}[!t]
\centering
\footnotesize
\setlength{\tabcolsep}{1.8pt}
\renewcommand{\arraystretch}{1.05}
\begin{tabular}{@{}lrrrr@{}}
\toprule
Predictors & $\rho\uparrow$ & MAE $\downarrow$
& RMSE $\downarrow$ & $Q^2\uparrow$ \\
\midrule
Training-mean baseline & -- & 8.46 & 10.40 & 0.000 \\
Foundational profile $F$ & 0.416 & 6.91 & 8.18 & 0.382 \\
Operation profile $O$ & 0.693 & 4.83 & 5.68 & 0.702 \\
Joint profile $F+O$ & \textbf{0.776} & \textbf{4.63} & \textbf{5.38} & \textbf{0.733} \\
\bottomrule
\end{tabular}
\caption{\textbf{Operation profiles improve model-family-held-out
prediction of Banking task reward.} All learned predictors use
$\alpha=10$; $\rho$ is Spearman correlation, errors are in reward
points, and $Q^2$ uses the fold-specific training-mean baseline. Arrows
mark the preferred direction; bold marks the best estimate.}
\label{tab:banking-reward-prediction}
\end{table}

At the model level, operation information corrects several overestimates
from the foundational profile, especially for Gemini 3.1 Pro and MiniMax
M3. The gain appears across most held-out families and under neighboring
regularization settings documented in
Appendix~\ref{app:alpha10-reference}. Family-block resampling places the
joint-over-foundational error reduction above zero. The joint and
operation-only predictors remain close, which attributes most of the gain
to operation information and a smaller complementary role to foundational
features. Pairwise model ordering follows the same progression from
foundational to operation and joint profiles.

Foundational profiles estimate expected execution before deployment;
operation profiles reveal service-specific tendencies after trajectories
are observed. Their error difference measures the value of observing
behavior directly, and the joint gain measures the remaining capability
signal. Operation measurements dominate in this cohort, with a smaller
complementary correction from foundational features.

\subsection{Further Analysis}
\label{sec:more-analysis}

\Needspace{7\baselineskip}
\resultbox{Takeaway}{Capability realization gap rankings persist across disjoint task subsets.}

\paragraph{Cross-Task Stability of Capability Realization Gaps}
\label{sec:results-gap-stability}
Overall gap rankings remain stable across repeated fixed task-group splits.
Each side of a split is scored and predicted independently, with scenario
groups kept intact. Progress Monitoring gaps are especially stable;
Recovery varies more across task composition. Appendix
Table~\ref{tab:gap-cross-task-stability} reports the primary split and
Appendix~\ref{app:gap-split-results} summarizes the sensitivity analysis.
The recurring rankings show that some model-level deviations from
foundational predictions persist across tasks.

Each subset re-estimates expected operation performance. Stable rankings
therefore reflect recurring execution deviations across independent fits.
Differences by operation expose sensitivity to task composition; Recovery
depends on an observable failure and a chance to respond.

\Needspace{7\baselineskip}
\resultbox{Takeaway}{Local edits produce occasional execution changes with limited repeatability.}

\paragraph{Local Content Interventions}
\label{sec:results-interventions}
Prediction establishes an association between operation performance and
task outcomes; interventions test whether local operation content changes
task progress. We restore checkpoints from DeepSeek V4 Flash, MiniMax M3,
and GLM-5.2, then compare original, edited, and control branches. Most
intervention strata show no mean score change. A tool-identifier repair
succeeds in one repetition and fails to reproduce; a planning-text repair
leaves the submitted call and final score unchanged. Local correctness is
therefore one factor within the broader trajectory state. Appendix
\ref{app:content-intervention-protocol} specifies the protocol, and
Appendix~\ref{app:content-intervention-results} reports aggregate,
example-level, and model-level results.

The control branch separates the edit from ordinary rerun variation.
Divergence after the same repair shows that later tool results and decisions
mediate its effect. The planning edit changes internal text and leaves the
submitted action fixed. A local improvement can therefore leave the final
task reward unchanged.

\Needspace{7\baselineskip}
\resultbox{Takeaway}{Shared-initial-state experiments reproduce differences in constraint execution across repeated continuations.}

\paragraph{Repeated Execution Differences from Shared Initial States}
\label{sec:results-shared-initial-states}
Differences in constraint execution recur when models continue from
shared initial states. Gemini 3.1 Pro, GLM-5.2, and Claude Fable 5
continue from matched dialogue histories, environment states, and tool
conditions under natural continuation and a generic checking prompt.
Scoring distinguishes policy violations, completed compliant operations,
and unresolved behaviors such as policy lookup or requests for
confirmation. Gemini 3.1 Pro shows the most repeated violations, GLM-5.2
shows fewer, and Claude Fable 5 shows only occasional recurrence.
Table~\ref{tab:shared-state-summary} summarizes this ordering. It appears
under both prompt conditions, and later steps sometimes reverse an
initially compliant action. The pattern locates model-specific execution
differences downstream of a shared history and tool state.
Appendix~\ref{app:shared-state-protocol} gives the protocol;
Appendix~\ref{app:shared-state-results} provides the checkpoint summary
and representative cases.

Matched checkpoints hold the prior dialogue, database, available tools, and
user goal constant. Recurring differences therefore emerge during the new
continuation. Persistence across natural and checking prompts further shows
that a generic request to inspect constraints leaves the execution pattern
largely intact. The later-step reversals reveal another source of variation:
an initially compliant action can be followed by a conflicting decision.
Repeated continuations expose step-level compliance changes hidden by one
endpoint reward.

\section{Conclusion}
\label{sec:conclusion}

We introduce a framework linking foundational capabilities, agentic
operations, and task outcomes through capability realization gaps. In
Banking, foundational profiles explain part of operation performance, with
persistent model-level gaps across task subsets. Operation profiles add
task-score information, and shared-initial-state experiments reveal
repeatable execution differences. Local repairs produce narrower,
case-dependent effects. Agent evaluation should measure capabilities and
their reliable deployment during interaction.

\section*{Limitations}
\label{sec:limitations}

The study evaluates one domain, Banking. Future work should validate the
framework in additional domains and agent environments, including whether
the operation definitions, capability realization gaps, and predictive
gains transfer across task structures.

% =========================
% References
% =========================

\bibliography{Reference}

% =========================
% Appendix
% =========================

\clearpage
\appendix

% Keep the appendix, including its contents page, in the ACL two-column layout.
\section*{Appendix Contents}
\phantomsection
\label{app:overview}

\begingroup
\footnotesize
\newcommand{\appsectionline}[1]{%
  \par\smallskip\noindent
  \hyperref[#1]{\textbf{\ref*{#1}\quad\nameref*{#1}}}%
  \dotfill\pageref{#1}\par}
\newcommand{\appsubsectionline}[1]{%
  \noindent\hspace{1.1em}%
  \hyperref[#1]{\ref*{#1}\quad\nameref*{#1}}%
  \dotfill\pageref{#1}\par}

\appsectionline{app:benchmark-details}
\appsubsectionline{app:agentic-benchmark-snapshot}
\appsubsectionline{app:trajectory-annotation-protocol}
\appsectionline{app:experimental-details}
\appsubsectionline{app:model-configurations}
\appsubsectionline{app:foundational-evaluation-config}
\appsubsectionline{app:harness-config}
\appsubsectionline{app:prediction-protocol}
\appsubsectionline{app:gap-stability-protocol}
\appsubsectionline{app:content-intervention-protocol}
\appsubsectionline{app:shared-state-protocol}
\appsectionline{app:prompts}
\appsubsectionline{app:self-monitoring-prompts}
\appsubsectionline{app:annotation-prompts}
\appsectionline{app:additional-results}
\appsubsectionline{app:alpha10-reference}
\appsubsectionline{app:historical-nested-protocol}
\appsubsectionline{app:ce-recovery-sensitivity}
\appsubsectionline{app:gap-split-results}
\appsubsectionline{app:content-intervention-results}
\appsubsectionline{app:shared-state-results}
\appsectionline{app:data-statement}
\appsectionline{app:notation}
\endgroup

\section{Benchmark Details}
\label{app:benchmark-details}

\subsection{Agentic Benchmark Snapshot}
\label{app:agentic-benchmark-snapshot}

The Banking corpus contains 1,451 trajectories from fifteen model services, with 96 or 97 trajectories per model. Each record includes user and agent messages, tool calls, tool results, and the native task reward. Analyses use the task definitions and policies saved with the historical runs. The common-task subset contains ninety-three tasks per model, or 1,395 trajectories.

\subsection{Trajectory Annotation Protocol}
\label{app:trajectory-annotation-protocol}

Task contracts specify the user goal, relevant entities, constraints, required actions, and valid terminal states. Annotations use the complete chronological trajectory, with model identity and task reward hidden from the annotation input. User and tool messages supply evidence; labels evaluate the agent's decisions and actions. A record can span several steps. The scoring evidence and label rule for each operation are given in Section~\ref{sec:operation-profile} and Table~\ref{tab:operation-annotation-rules}. Missing required actions receive zero; simulator deviations are recorded separately. The annotation pipeline automatically accepted 1,439 trajectories; twelve additional trajectories were adjudicated by an AI assistant. The AI-assisted adjudications fall outside independent human double annotation.

\section{Experimental Details}
\label{app:experimental-details}

\subsection{Model and Inference Configurations}
\label{app:model-configurations}

Model services are listed in Table~\ref{tab:model-profiles}. Foundational scores combine public results and local measurements; public scores retain their source-specific inference settings and evaluation cohorts. Model-level sources, reasoning settings, and effective denominators are recorded in the supplementary materials.

\subsection{Foundational Evaluation Configuration}
\label{app:foundational-evaluation-config}

Table~\ref{tab:foundational-battery} in Section~\ref{sec:experimental-setup} lists the measurement definitions and reference dataset sizes. Effective cohorts can differ from these reference sizes. In particular, some local Self-Monitoring results use 148 GPQA items, and valid-response estimates retain their smaller denominators. Self-Monitoring uses an answer followed by a confidence report, without correctness feedback.

\subsection{Agent Harness Configuration}
\label{app:harness-config}

Historical Banking runs use the saved task files, tool interfaces, policies, and native reward evaluation. DeepSeek-V4-Pro supplies user-simulator responses with temperature zero. Checkpoint experiments restore the saved database, dialogue history, tool availability, and knowledge-base state. Local content interventions preserve the source agent/user configuration and remaining budget. Shared-initial-state experiments retain each service's native reasoning settings and add no new user messages or confirmations.

\subsection{Prediction and Uncertainty Protocol}
\label{app:prediction-protocol}

Each outer fold holds out an entire model family. Ridge predictors use training-fold standardization, an unpenalized intercept, equal model weights, and unclipped predictions. The common $\alpha=10$ follows an exploratory cohort-level penalty scan; Appendix~\ref{app:alpha10-reference} retains neighboring settings and the nested-selection result. Reward is multiplied by 100. Fixed-prediction error intervals resample the eight model families 100,000 times while keeping the fitted predictions fixed.

\subsection{Cross-Task Stability Checks}
\label{app:gap-stability-protocol}

The ninety-three common tasks form seventeen scenario groups. The primary split contains 37 tasks in seven groups and 56 tasks in ten groups; no group crosses the split. Each side independently aggregates operation scores and fits model-family-held-out predictors. Intervals use 1,000 within-side scenario-group resamples, refitting predictors each time. Twenty additional fixed group splits assess partition sensitivity.

\subsection{Local Content Intervention Protocol}
\label{app:content-intervention-protocol}

The main cohort contains 38 frozen cases from thirteen tasks and three models, covering ST, PL, AS, and TM. Each case plans three repetitions of original ($H_0$), edited ($M$), and control ($C$) branches from the same restored checkpoint. Text controls preserve meaning; tool-call controls preserve the original call. Editing text accompanying a pending call leaves that call unchanged. There are 98 complete groups out of 114 planned groups; incomplete groups are recorded as missing. Damage effects are $H_0-M$ and repair effects are $M-H_0$, using native task reward. Repetitions are averaged within cases, cases within tasks, and tasks equally within each stratum. Stratum estimates are descriptive because each has fewer than five contributing tasks.

\subsection{Shared-Initial-State Experiment Protocol}
\label{app:shared-state-protocol}

Gemini 3.1 Pro, GLM-5.2, and Claude Fable 5 continue from historical action-trigger checkpoints, including previously inspected cases. The four rounds are summarized in Table~\ref{tab:shared-state-rounds}. There are eighteen distinct task IDs and 195 continuations, of which 193 are scoreable; two GLM outputs in task 040 are missing. The final round repeats task 089 after its earlier result.

\begin{table}[!htbp]
\centering
\small
\renewcommand{\arraystretch}{1.12}
\caption{\textbf{Coverage of the shared-initial-state experiments.}
States are restored checkpoints, repeats are continuations per
condition, and steps count assistant decisions. N denotes natural
continuation and I the generic checking prompt.}
\label{tab:shared-state-rounds}
\setlength{\tabcolsep}{3pt}
\begin{tabular}{@{}lrrrr@{}}
\toprule
\textbf{Round} & \textbf{States} & \textbf{Repeats} & \textbf{Steps} & \textbf{Runs} \\
\midrule
Initial (N/I) & 4 & 3 & 1 & 72 \\
Expansion (N) & 8 & 3 & 1 & 72 \\
Short continuation (N) & 4 & 2 & $\leq3$ & 24 \\
Final (N) & 3 & 3 & $\leq3$ & 27 \\
\bottomrule
\end{tabular}
\end{table}

Scoring separates local progress, violations, and unresolved confirmation or policy lookup. Later rounds also check compliance at the end of the continuation. The initial round uses two AI ratings with model metadata hidden; later rounds use AI-assistant review of histories, tool receipts, and state changes. Checkpoint-level summaries count each task ID once per model; repeated violations require at least two repetitions within one round and prompt condition.

\Needspace{26\baselineskip}
\section{Prompts}
\label{app:prompts}

Natural continuations retain the restored instructions, dialogue
history, and tools. The checking condition appends the following
identical instruction to the existing system prompt:

\begin{agentpromptbox}{\thesection: Checking instruction}

% Target badge and condition
\tcbox[
  on line,
  colback=promptbadge,
  colframe=promptborder,
  boxrule=0.4pt,
  arc=2mm,
  boxsep=0pt,
  left=1.5mm,
  right=1.5mm,
  top=0.6mm,
  bottom=0.6mm
]{%
  \footnotesize
  \color{promptnavy}
  \textbf{Target:} System prompt
}
\hfill
{\footnotesize\color{promptnavy}Checking condition}

\par\medskip
{\color{promptnavy}\bfseries INSTRUCTION}
\par\smallskip

Before acting, identify the user's currently authorized goal, the
applicable requirements, and any evidence still needed. Then take
the smallest lawful next step. Do not infer additional authorization
or claim a completed action without an earlier supporting result.
Use the normal tools and response format.

% Dashed separator
\tcblower

% These notes describe the protocol; they are not appended to the prompt.
{\color{promptnavy}\bfseries PROTOCOL NOTES}
\par\medskip

\textbf{Scope.}
The added instruction contains no task-specific answer or new
authorization.

\par\medskip
\textbf{Local edits.}
Local content interventions edit only the selected text or tool-call
field; the remaining restored prompt is unchanged.

\end{agentpromptbox}
\subsection{Self-Monitoring Prompts}
\label{app:self-monitoring-prompts}

The following templates are used in the local two-stage measurement. Stage 2 receives the question and the model's original answer, without a reference answer or correctness feedback. Bracketed fields below stand for the question content.

\begin{agentpromptbox}{Self-Monitoring: answer and confidence}
\textbf{Stage 1: GPQA Diamond}
\par\smallskip
Answer the following multiple-choice question.
Return only a JSON object in the form \texttt{\{"answer": "A"\}}, where the value is one
of A, B, C, or D. Do not include confidence or any additional text.
\par\smallskip
QUESTION:\par
[Question and options]

\par\medskip
\textbf{Stage 1: SimpleQA}
\par\smallskip
Answer the following fact-seeking question as concisely as possible.
Return only a JSON object in the form \texttt{\{"answer": "..."\}}.
Do not include confidence or any additional text.
\par\smallskip
QUESTION:\par
[Question]

\tcblower
\textbf{Stage 2: both datasets}
\par\smallskip
Estimate the probability that your previous answer to this question is fully
correct. Do not solve the question again, critique the answer, or provide an
explanation. Return only a JSON object in the following form:\par
\texttt{\{"p\_correct": 0.0\}}
\par\smallskip
\texttt{p\_correct} must be a number between 0 and 1 inclusive.
\end{agentpromptbox}

\subsection{Automatic Annotation Prompts}
\label{app:annotation-prompts}

The two-pass annotator first builds a task contract and then judges the trajectory against it. Below are verbatim excerpts from the contract prompt (v3.8.1) and the frozen scoring prompt (v3.8.9). The full templates are included in the accompanying prompt supplement. Scoring and aggregation follow Section~\ref{sec:operation-profile} and Appendix~\ref{app:trajectory-annotation-protocol}.

\begin{agentpromptbox}{Automatic annotation: prompt excerpts}
\textbf{Pass A: task contract}
\par\smallskip
You are Pass A of a blinded two-pass evaluator for agent trajectories. Build the frozen task
contract only. Do not score the agent and do not output any seven-dimension labels.

\tcblower
\textbf{Pass B: trajectory judgments}
\par\smallskip
Audit a blinded trajectory against fixed contract items, stage checks, and any supplied outcome
anchors. Return JSON only. Do not infer reward, model identity, prior labels, or a score distribution.
Keep supplied order. Local code derives all 0/0.5/1 dimension scores.
\end{agentpromptbox}

Pass A returns the task contract, stages, and source citations. Pass B receives the blinded episode, frozen contract, item targets, stage checks, termination state, and outcome anchors. Its JSON format is shown below with the illustrative values from the actual prompt. Statuses are \texttt{satisfied}, \texttt{partial}, \texttt{failed}, \texttt{omitted}, or \texttt{not\_exposed}; turn indices locate supporting evidence and gaps. The internal operation codes G/S/P/A/M/R/T correspond to GI/ST/PL/AS/PM/RC/TM.

{\scriptsize
\begin{verbatim}
{
  "protocol_version": "v3.8.9-outcome-anchors",
  "frozen_contract_sha256": "...",
  "contract_judgments": [{
    "item_id": "st1-ra1", "status": "failed",
    "positive_turns": [], "gap_turns": [4],
    "reason": "concise evidence"
  }],
  "stage_checks": [{
    "check_id": "st1:G", "status": "partial",
    "positive_turns": [2], "gap_turns": [8],
    "reason": "concise evidence"
  }],
  "issues": []
}
\end{verbatim}
}

\raggedbottom
\section{Additional Results}
\label{app:additional-results}

\subsection{Common-Penalty Reference and Sensitivity}
\label{app:alpha10-reference}
This section reports prediction diagnostics and sensitivity to ridge
regularization under the same model cohort and model-family-held-out
splits. Across the reported settings, the joint profile has lower MAE
than the foundational profile.

\noindent\begin{minipage}{\columnwidth}
\captionsetup{type=table}
\centering
\footnotesize
\caption{\textbf{Prediction performance under alternative ridge
penalties.} Fixed-penalty and nested-selection results use the same
model cohort and outer family splits. Macro denotes operation-macro MAE,
and $\rho$ denotes Spearman correlation for Banking reward.}
\label{tab:alpha10-neighborhood}
\setlength{\tabcolsep}{1.4pt}
\renewcommand{\arraystretch}{1.10}
\begin{tabular}{@{}>{\raggedright\arraybackslash}p{0.31\columnwidth}rrrrr@{}}
\toprule
\textbf{Setting} & \textbf{Macro} & $F$ & $O$ & $F+O$ & $F+O$ \\
 & MAE & MAE & MAE & MAE & $\rho$ \\
\midrule
Nested selection & 7.99 & 7.80 & 4.84 & 6.07 & 0.758 \\
Fixed $\alpha=5$ & 4.70 & 7.43 & 4.95 & 4.95 & 0.810 \\
Fixed $\alpha=10$ & 4.47 & 6.91 & 4.83 & 4.63 & 0.776 \\
Fixed $\alpha=15$ & 4.47 & 6.84 & 4.72 & 4.49 & 0.763 \\
Fixed $\alpha=20$ & 4.53 & 6.85 & 4.65 & 4.49 & 0.718 \\
Fixed $\alpha=30$ & 4.65 & 6.94 & 4.79 & 4.52 & 0.724 \\
\bottomrule
\end{tabular}
\end{minipage}

\par\medskip
\noindent\begin{minipage}{\columnwidth}
\captionsetup{type=table}
\centering
\footnotesize
\caption{\textbf{Operation-level prediction diagnostics under the common
ridge penalty.} Errors use 0--100 target scales at $\alpha=10$; Base is
the fold-specific training-mean MAE, $\rho$ is Spearman correlation, and
$\tau_b$ is Kendall's rank correlation.}
\label{tab:alpha10-operation-diagnostics}
\setlength{\tabcolsep}{2pt}
\renewcommand{\arraystretch}{1.10}
\begin{tabular}{@{}lrrrrrr@{}}
\toprule
\textbf{Target} & \textbf{MAE} & \textbf{Base} & \textbf{RMSE} & $Q^2$ & $\rho$ & $\tau_b$ \\
\midrule
GI & 4.28 & 5.76 & 5.74 & 0.288 & 0.150 & 0.143 \\
ST & 4.26 & 5.31 & 5.13 & 0.357 & 0.282 & 0.143 \\
PL & 4.01 & 5.21 & 4.69 & 0.411 & 0.125 & 0.105 \\
AS & 4.99 & 6.89 & 6.13 & 0.437 & 0.321 & 0.276 \\
PM & 6.46 & 7.18 & 7.60 & 0.260 & 0.159 & 0.096 \\
RC & 7.24 & 8.77 & 10.12 & 0.175 & 0.182 & 0.162 \\
TM & 5.36 & 5.79 & 6.30 & 0.151 & 0.102 & 0.096 \\
Op. macro & 4.47 & 5.92 & 5.63 & 0.373 & 0.168 & 0.105 \\
\bottomrule
\end{tabular}
\end{minipage}

\par\medskip
Across nearby penalties, the joint profile improves on foundational
features and remains close to the operation-only fit. Action Selection and
Planning are the most predictable operations; Recovery and Termination are
harder.

\begin{figure*}[t]
\centering
\includegraphics[width=0.80\textwidth]{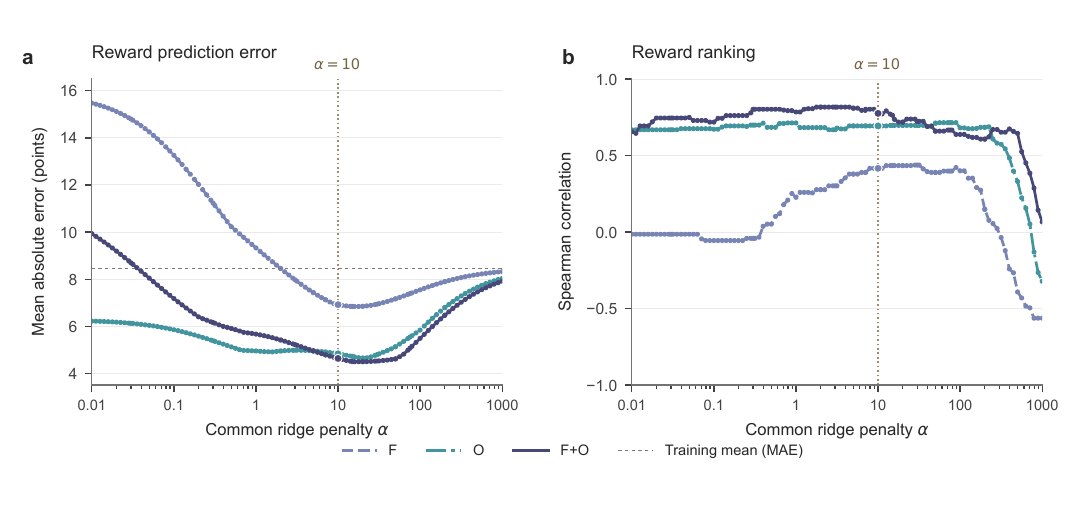}
\caption{\textbf{Sensitivity of Banking reward prediction to ridge
regularization.} Curves compare foundational ($F$), operation ($O$),
and joint ($F+O$) inputs using (a) MAE and (b) Spearman correlation.
The horizontal axis is logarithmic, and the vertical dashed line marks
the common setting $\alpha=10$.}
\label{fig:alpha10-regularization-sensitivity}
\end{figure*}

\begin{table*}[t]
\centering
\small
\setlength{\tabcolsep}{4pt}
\renewcommand{\arraystretch}{1.15}

\begin{tabularx}{\textwidth}{
  @{}l
  *{7}{>{\centering\arraybackslash}X}
  @{}
}
\toprule

\multirow{2}{*}{\textbf{Model}}
& \multicolumn{7}{c}{\textbf{Agentic Operation Residuals}} \\
\cmidrule(lr){2-8}

& \textbf{GI}
& \textbf{ST}
& \textbf{PL}
& \textbf{AS}
& \textbf{PM}
& \textbf{RC}
& \textbf{TM} \\

\midrule

% =========================
% OpenAI
% =========================

\modelicon{openai}GPT-5.6 Sol
& $+1.13$ & $-2.74$ & $-0.78$ & $-6.26$
& $-4.70$ & $+5.11$ & $-6.31$ \\

\modelicon{openai}GPT-5.6 Luna
& $+0.46$ & $-0.76$ & $-2.80$ & $+3.10$
& $+1.93$ & $-3.48$ & $+3.92$ \\

\modelicon{openai}GPT-5.5
& $+0.19$ & $-0.78$ & $+4.62$ & $+2.10$
& $-3.00$ & $-14.60$ & $-0.86$ \\

% =========================
% Anthropic
% =========================

\arrayrulecolor{black!35}\midrule[0.3pt]\arrayrulecolor{black}
\modelicon{anthropic}Claude Opus 5
& $+4.28$ & $+4.41$ & $+5.91$ & $+7.33$
& $+7.56$ & $+19.63$ & $+4.23$ \\

\modelicon{anthropic}Claude Fable 5
& $+8.59$ & $+9.05$ & $+3.64$ & $+6.12$
& $+7.66$ & $+21.38$ & $+8.12$ \\

\modelicon{anthropic}Claude Sonnet 5
& $+5.60$ & $+3.07$ & $+4.79$ & $+8.98$
& $+2.93$ & $+9.84$ & $+4.49$ \\

% =========================
% Google
% =========================

\arrayrulecolor{black!35}\midrule[0.3pt]\arrayrulecolor{black}
\modelicon{google}Gemini 3.1 Pro
& $-15.51$ & $-9.57$ & $-10.12$ & $-13.01$
& $-16.97$ & $-16.31$ & $-14.42$ \\

\modelicon{google}Gemini 3.6 Flash
& $-2.15$ & $+3.06$ & $-1.18$ & $+1.01$
& $-2.90$ & $+1.78$ & $-0.90$ \\

% =========================
% Moonshot AI
% =========================

\arrayrulecolor{black!35}\midrule[0.3pt]\arrayrulecolor{black}
\modelicon{moonshot}Kimi K3
& $-0.15$ & $-2.00$ & $-6.36$ & $-0.53$
& $+4.88$ & $-6.77$ & $+1.27$ \\

\modelicon{moonshot}Kimi K2.6
& $-2.27$ & $-8.79$ & $-2.41$ & $-2.37$
& $-2.36$ & $-0.30$ & $-4.66$ \\

% =========================
% DeepSeek
% =========================

\arrayrulecolor{black!35}\midrule[0.3pt]\arrayrulecolor{black}
\modelicon{deepseek}DeepSeek V4 Pro
& $+3.91$ & $+5.14$ & $+2.86$ & $-1.27$
& $+5.52$ & $+0.92$ & $+4.55$ \\

\modelicon{deepseek}DeepSeek V4 Flash
& $-6.03$ & $-3.96$ & $-2.67$ & $-8.48$
& $-7.41$ & $+4.71$ & $-5.62$ \\

% =========================
% Z.ai
% =========================

\arrayrulecolor{black!35}\midrule[0.3pt]\arrayrulecolor{black}
\modelicon{zai}GLM-5.2
& $+4.36$ & $+6.54$ & $+7.02$ & $+2.48$
& $+7.05$ & $-0.75$ & $+5.83$ \\

% =========================
% Qwen
% =========================

\arrayrulecolor{black!35}\midrule[0.3pt]\arrayrulecolor{black}
\modelicon{qwen}Qwen3.8-Max
& $+5.68$ & $+1.46$ & $+3.51$ & $+8.41$
& $+9.78$ & $+0.38$ & $+7.84$ \\

% =========================
% MiniMax
% =========================

\arrayrulecolor{black!35}\midrule[0.3pt]\arrayrulecolor{black}
\modelicon{minimax}MiniMax M3
& $-3.94$ & $+2.64$ & $+1.46$ & $-3.35$
& $-12.32$ & $+2.66$ & $-7.41$ \\

\bottomrule
\end{tabularx}

\caption{\textbf{Operation-level capability realization gaps for every
model service.} Values are observed operation scores minus
model-family-held-out predictions at $\alpha=10$; they therefore differ
from the overall gap produced by the separate macro predictor. Provider
icons and separators identify model families.}

\label{tab:alpha10-operation-residuals}

\end{table*}

\begin{figure*}[t]
\centering
\includegraphics[width=0.68\textwidth]{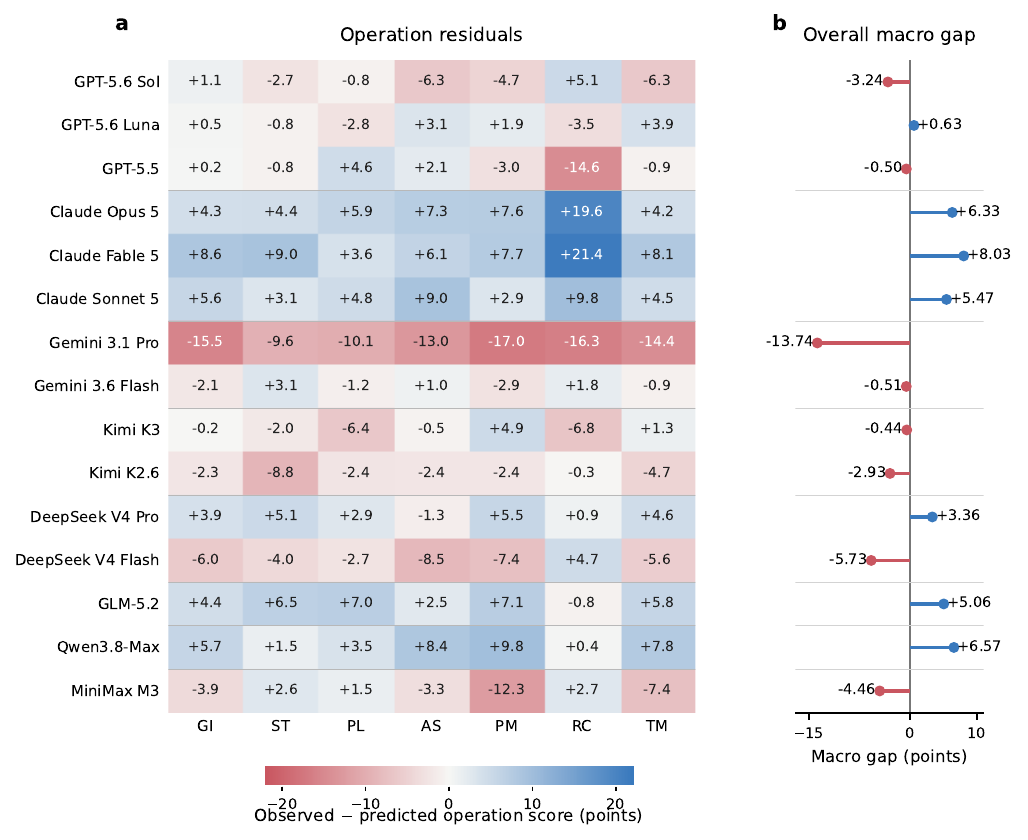}
\caption{\textbf{Capability realization gaps by model and operation.}
(a) Observed operation scores minus model-family-held-out predictions
from foundational profiles at $\alpha=10$. (b) Gaps from a separate
predictor of the overall operation score reported in
Table~\ref{tab:overall-operation-gap}. Blue indicates execution above
prediction and red below prediction; exact values appear in
Table~\ref{tab:alpha10-operation-residuals}.}
\label{fig:operation-residuals}
\end{figure*}

\subsection{Nested-Selection Protocol}
\label{app:historical-nested-protocol}
The nested-selection protocol selects $\alpha$ from $\{0.01,0.1,1,10,100,1000\}$ by inner family holdouts, minimizing equally weighted family-level mean squared error. Standardization and selection use only the outer training fold. Selection is repeated separately for each target. The main results use the common $\alpha=10$ reference. Source records and the full scan are provided in the supplementary materials.

\subsection{CE--Recovery Association and Sensitivity}
\label{app:ce-recovery-sensitivity}
\begin{figure*}[t]
\centering
\includegraphics[width=0.76\textwidth]{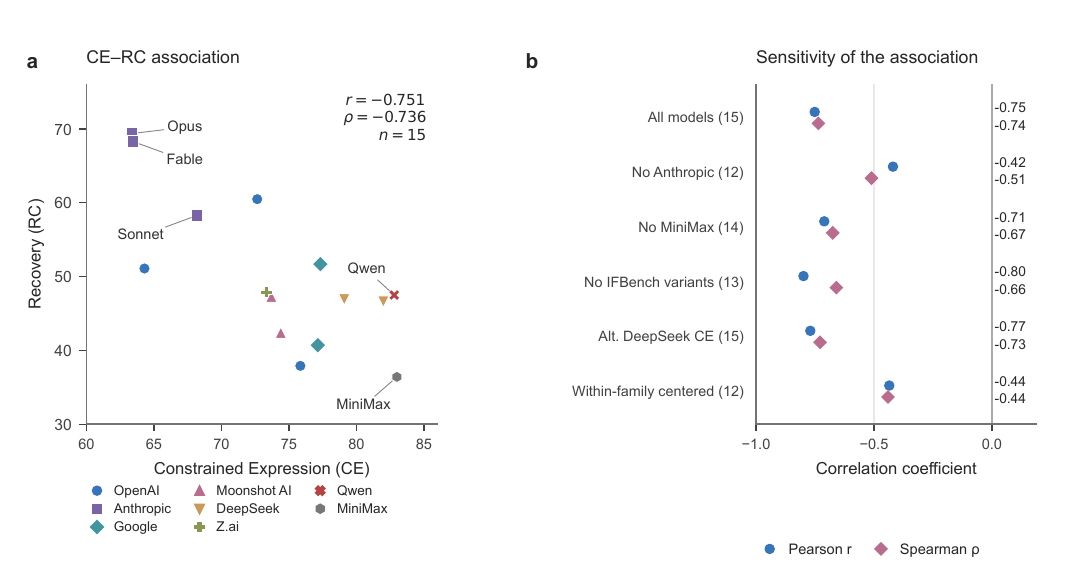}
\caption{\textbf{Association between Constrained Expression and
Recovery across analysis specifications.} (a) Raw CE and RC scores for
15 models. (b) Pearson and Spearman correlations under six sensitivity
specifications; parentheses give the number of included models.}
\label{fig:ce-recovery-sensitivity}
\end{figure*}

Constrained Expression is the only foundational dimension whose
associations with all seven operations are negative
in Figure~\ref{fig:capability-operation-mapping}a. Its strongest
association is with Recovery, with Pearson $r=-0.751$ and Spearman
$\rho=-0.736$ across all 15 models. Claude Opus 5 and Claude Fable 5
combine the lowest Constrained Expression scores with the highest
Recovery scores. The association remains negative after removing all
Anthropic models, which gives $r=-0.42$, and after centering scores
within each family, which gives $r=-0.44$. Three further specifications
show the same direction. This descriptive cohort-level pattern separates
format-constrained instruction following from self-repair during
interaction.
Figure~\ref{fig:ce-recovery-sensitivity} reports the raw scores and all
six sensitivity specifications.

\subsection{Cross-Task Split Sensitivity}
\label{app:gap-split-results}

\noindent\begin{minipage}{\columnwidth}
\captionsetup{type=table}
\centering
\small
\caption{\textbf{Capability realization gaps are partly stable across
the primary pair of disjoint task subsets.} Spearman correlations compare
gaps estimated independently on the two subsets; 95\% intervals are
obtained by resampling scenario groups within each subset.}
\label{tab:gap-cross-task-stability}
\setlength{\tabcolsep}{6pt}
\begin{tabular}{@{}lrc@{}}
\toprule
\textbf{Operation} & $\boldsymbol{\rho}$ & \textbf{95\% interval} \\
\midrule
GI & 0.461 & $[0.021,\,0.657]$ \\
ST & 0.311 & $[-0.268,\,0.575]$ \\
PL & 0.421 & $[-0.171,\,0.607]$ \\
AS & 0.554 & $[-0.164,\,0.750]$ \\
PM & 0.800 & $[0.403,\,0.907]$ \\
RC & 0.011 & $[-0.436,\,0.482]$ \\
TM & 0.650 & $[0.082,\,0.786]$ \\
Overall & 0.800 & $[0.300,\,0.836]$ \\
\bottomrule
\end{tabular}
\end{minipage}

\par\medskip
\noindent\begin{minipage}{\columnwidth}
\captionsetup{type=table}
\centering
\footnotesize
\renewcommand{\arraystretch}{1.10}
\caption{\textbf{Capability realization gap stability across 20
additional task-group splits.} Entries report the median and range of
Spearman rank correlations between the two disjoint sides of each fixed
split.}
\label{tab:gap-split-sensitivity}
\begin{tabular}{@{}lrc@{}}
\toprule
\textbf{Operation} & Median $\rho$ & \textbf{Range} \\
\midrule
GI & 0.552 & $[0.339,\,0.750]$ \\
ST & 0.180 & $[-0.164,\,0.764]$ \\
PL & 0.409 & $[-0.364,\,0.775]$ \\
AS & 0.425 & $[0.104,\,0.893]$ \\
PM & 0.761 & $[0.471,\,0.886]$ \\
RC & 0.202 & $[-0.111,\,0.561]$ \\
TM & 0.645 & $[0.332,\,0.821]$ \\
Overall & 0.759 & $[0.554,\,0.861]$ \\
\bottomrule
\end{tabular}
\end{minipage}

Both split analyses give the same ordering: Progress Monitoring and the
overall gap are the most stable, while Recovery is the most sensitive to
task composition.

\subsection{Local Content Intervention Results}
\label{app:content-intervention-results}

\noindent\begin{minipage}{\columnwidth}
\captionsetup{type=table}
\centering
\footnotesize
\renewcommand{\arraystretch}{1.10}
\caption{\textbf{Model-level outcomes of the local content intervention
experiment.} Groups are complete/planned three-branch repetitions.
$D$ is the direction-adjusted original--edited reward difference,
$K=C-H_0$, and $E=C-M$ for damage or $E=M-C$ for repair. Effects are
complete-group-weighted descriptive means on the 0--1 reward scale.}
\label{tab:model-content-interventions}
\setlength{\tabcolsep}{3pt}
\begin{tabular}{@{}lrrrr@{}}
\toprule
\textbf{Model} & \textbf{Groups} & $D$ & $K$ & $E$ \\
\midrule
DeepSeek V4 Flash & 30/39 & $-0.030$ & $-0.004$ & $-0.033$ \\
GLM-5.2 & 29/33 & $-0.069$ & $+0.069$ & $0.000$ \\
MiniMax M3 & 39/42 & $+0.051$ & $+0.026$ & $+0.077$ \\
\bottomrule
\end{tabular}
\end{minipage}

\par\medskip
\noindent\begin{minipage}{\columnwidth}
\captionsetup{type=table}
\centering
\footnotesize
\caption{\textbf{Aggregate outcomes of the local content intervention
experiment.} Counts summarize complete three-branch groups and strata;
repair rows list final rewards for the original, edited, and control
branches in chronological order.}
\label{tab:local-content-interventions}
\setlength{\tabcolsep}{3pt}
\renewcommand{\arraystretch}{1.10}
\begin{tabularx}{\columnwidth}{@{}>{\raggedright\arraybackslash}Xl@{}}
\toprule
\textbf{Measure} & \textbf{Result} \\
\midrule
Complete three-branch groups & 98/114 \\
Strata with zero mean effect & 12/17 \\
\midrule
Tool-identifier repair & $(0,1,0)$; $(0,0,0)$ \\
Planning-text repair & $(0,0,0)$; $(0,0,0)$ \\
\bottomrule
\end{tabularx}
\end{minipage}

\par\medskip
Effects remain small and model dependent. MiniMax M3 has positive
contrasts, DeepSeek V4 Flash has small negative contrasts, and only one
tool-identifier repetition changes reward.

\subsection{Shared-Initial-State Experiment Results}
\label{app:shared-state-results}

The checkpoint-level summary appears in
Table~\ref{tab:shared-state-summary}. At the stolen-card checkpoint,
Gemini 3.1 Pro permanently blocks the card without the required
verification in all three repetitions under both natural continuation
and the checking prompt. GLM-5.2 and Claude Fable 5 avoid this violation
and leave the protective action incomplete in the first step. At the
temporary ATM-limit checkpoint, Gemini first increases the eligible
account's limit and then also increases an ineligible account's limit;
this recurs in both the initial short continuations and the added
repetitions. GLM and Fable avoid the corresponding invalid increase.
The latter case shows how a later action can undermine a
correct first step.

\noindent\begin{minipage}{\columnwidth}
\captionsetup{type=table}
\centering
\footnotesize
\caption{\textbf{Repeated violations from shared initial states.}
``Any'' denotes at least one violation, while ``repeated'' requires
violations in at least two repetitions within the same round and prompt
condition. Each task checkpoint is counted once per model.}
\label{tab:shared-state-summary}
\setlength{\tabcolsep}{3pt}
\renewcommand{\arraystretch}{1.10}
\begin{tabularx}{\columnwidth}{@{}>{\raggedright\arraybackslash}Xrr@{}}
\toprule
\textbf{Model} & \textbf{Any violation} & \textbf{Repeated violation} \\
\midrule
Gemini 3.1 Pro & 33.3\% (6/18) & 22.2\% (4/18) \\
GLM-5.2 & 16.7\% (3/18) & 11.1\% (2/18) \\
Claude Fable 5 & 5.6\% (1/18) & 5.6\% (1/18) \\
\bottomrule
\end{tabularx}
\end{minipage}

\Needspace{8\baselineskip}
\section{Data Statement and License}
\label{app:data-statement}

Foundational capability scores combine cited public evaluations with local
measurements. Banking trajectories come from model services in the
simulated Banking environment built on
$\tau^2$-Bench~\citep{barres2025tau2bench} and include messages, tool
interactions, and environment rewards. Supplementary archives provide
model profiles, held-out predictions, sensitivity results, and prompt
templates. The $\tau^2$-Bench software uses the MIT
license.\footnote{\url{https://github.com/sierra-research/tau2-bench/blob/main/LICENSE}}
Third-party benchmark data retain their original licenses and access terms.

\clearpage
\twocolumn[{%
\section{Notation}
\label{app:notation}

Table~\ref{tab:notation} summarizes the notation used throughout the paper. Vectors are column vectors unless stated otherwise.

\begin{center}
\captionsetup{type=table}
\footnotesize
\setlength{\tabcolsep}{3pt}
\renewcommand{\arraystretch}{1.10}
\begin{minipage}[t]{0.485\textwidth}
\vspace{0pt}
\begin{tabularx}{\linewidth}{@{}>{\raggedright\arraybackslash}p{0.18\linewidth}>{\raggedright\arraybackslash}p{0.23\linewidth}>{\raggedright\arraybackslash}X@{}}
\toprule
\textbf{Symbol} & \textbf{Type} & \textbf{Description} \\
\midrule
\rowcolor{black!6}\multicolumn{3}{@{}l}{\textit{Models and profiles}} \\
$\mathcal{M}$ & Set & Evaluated model services. \\
$m$ & Index & Model service under a fixed inference protocol. \\
$f$ & Mapping & Maps a model to its family; $f(m)$ is held out with $m$. \\
$o$ & Index & One of GI, ST, PL, AS, PM, RC, and TM. \\
$F_m=\Theta_m$ & Vector in $\mathbb{R}^7$ & Foundational profile of model $m$. \\
$O_m$ & Vector in $\mathbb{R}^7$ & Agentic operation profile of model $m$. \\
$O_{m,o}$ & Scalar & Score of model $m$ on operation $o$. \\
$A_m$ & Scalar & Overall operation score of model $m$. \\
$Y_m$ & Scalar target & Mean native Banking reward of model $m$ on a 0--100 scale. \\
$F,O,F+O$ & Feature sets & Foundational, operation, and joint predictor inputs. \\
\midrule
\rowcolor{black!6}\multicolumn{3}{@{}l}{\textit{Prediction}} \\
$\mathcal{I}_{\mathrm{train}}$ & Index set & Models in the current outer training fold. \\
$i$ & Index & Training model within an outer fold. \\
$y_i$ & Scalar & Target value for training model $i$. \\
$z_i$ & Vector & Standardized feature vector for model $i$. \\
$b$ & Scalar parameter & Unpenalized ridge intercept. \\
$\beta$ & Parameter vector & Ridge coefficients in Eq.~\ref{eq:ridge-objective}. \\
\bottomrule
\end{tabularx}
\end{minipage}\hfill
\begin{minipage}[t]{0.485\textwidth}
\vspace{0pt}
\begin{tabularx}{\linewidth}{@{}>{\raggedright\arraybackslash}p{0.18\linewidth}>{\raggedright\arraybackslash}p{0.23\linewidth}>{\raggedright\arraybackslash}X@{}}
\toprule
\textbf{Symbol} & \textbf{Type} & \textbf{Description} \\
\midrule
\rowcolor{black!6}\multicolumn{3}{@{}l}{\textit{Prediction (continued)}} \\
$\alpha$ & Scalar hyperparameter & Ridge penalty; the common setting is 10. \\
$t$ & Scalar target & Generic prediction target. \\
$\widehat{t}_m^{(-f(m))}$ & Scalar prediction & Prediction for $m$ trained without family $f(m)$. \\
$\overline{t}_{\mathrm{train}(m)}$ & Scalar baseline & Training-fold mean for the fold holding out $f(m)$. \\
\midrule
\rowcolor{black!6}\multicolumn{3}{@{}l}{\textit{Capability realization gaps and metrics}} \\
$r_{m,o}$ & Scalar residual & $O_{m,o}-\widehat O_{m,o}^{(-f(m))}$. \\
$\Delta_m^{\mathrm{op}}$ & Scalar residual & $A_m-\widehat A_m^{(-f(m))}$. \\
$Q^2(t)$ & Scalar metric & Held-out skill relative to the training mean. \\
$r,\rho,\tau_b$ & Scalar metrics & Pearson, Spearman, and Kendall correlations. \\
MAE, RMSE & Scalar metrics & Mean absolute and root mean squared error. \\
\midrule
\rowcolor{black!6}\multicolumn{3}{@{}l}{\textit{Interventions and shared initial states}} \\
$H_0,M,C$ & Branches & Original, locally edited, and control branches. \\
$D,K,E$ & Scalar contrasts & Direction-adjusted edit, control, and edit-control effects. \\
N, I & Conditions & Natural continuation and generic checking prompt. \\
\bottomrule
\end{tabularx}
\end{minipage}
\caption{\textbf{Mathematical objects used in the framework and
experiments.} The Type column distinguishes indices, sets, mappings,
vectors, scalar quantities, model parameters, metrics, experimental
branches, and conditions.}
\label{tab:notation}
\end{center}
}]

\end{document}